\documentclass{article}
\usepackage{PRIMEarxiv}

\usepackage[utf8]{inputenc}
\usepackage{booktabs}

\usepackage{mathtools}
\usepackage{bbm}

\usepackage{amsfonts}
\usepackage{amsmath}
\usepackage{comment}
\usepackage{multirow}
\usepackage{xcolor}

\newcommand{\script[1]}{{\mathcal{#1}}}

\newcommand{\scriptX}{{\mathcal{X}}}
\newcommand{\scriptC}{{\mathcal{C}}}
\newcommand{\scriptD}{{\mathcal{D}}}

\newcommand{\seenC}{{\mathcal{C}_s}}
\newcommand{\unseenC}{{\mathcal{C}_u}}

\newcommand{\Reals}{{\mathbb{R}}}

\newcommand{\colorblue[1]}{{\color{blue}{#1}}}
\newcommand{\colorred[1]}{{\color{red}{#1}}}
\usepackage{cite}
 
\usepackage{xr-hyper}
\makeatletter
\newcommand*{\addFileDependency}[1]{
  \typeout{(#1)}
  \@addtofilelist{#1}
  \IfFileExists{#1}{}{\typeout{No file #1.}}
}
\makeatother

\newcommand*{\myexternaldocument}[1]{%
    \externaldocument{#1}%
    \addFileDependency{#1.tex}%
    \addFileDependency{#1.aux}%
}

\myexternaldocument{supp_mat}

\usepackage{hyperref}       

\usepackage{graphicx}
\usepackage{caption}
\usepackage{subcaption}

\graphicspath{ {./figures/} }

\title{Using Fictitious Class Representations to Boost Discriminative Zero-Shot Learners}
\author{
  Mohammed Dabbah \\
  Department of Computer Science\\
  Technion – Israel Institute of Technology\\
  Haifa, Israel \\
  \texttt{mdabbah@campus.technion.ac.il} \\

  \And
    Ran El-Yaniv \\
  Department of Computer Science\\
  Technion – Israel Institute of Technology\\
  Haifa, Israel \\
  \texttt{rani@cs.technion.ac.il} \\
}
\date{April 2021}

\begin{document}

\maketitle

\begin{abstract}
Focusing on discriminative zero-shot learning, in this work we introduce a novel mechanism that dynamically augments during training the set of seen classes to produce additional fictitious classes. These  fictitious classes diminish the model's tendency to fixate during training on attribute correlations that appear in the training set but will not appear in newly exposed classes. The proposed model is tested within the two formulations of the zero-shot learning framework; namely, generalized zero-shot learning (GZSL) and classical zero-shot learning (CZSL). Our model improves the state-of-the-art performance on the CUB dataset and reaches comparable results on the other common datasets, AWA2 and SUN. We investigate the strengths and weaknesses of our method, including the effects of catastrophic forgetting when training an end-to-end zero-shot model.
\end{abstract}

\section{Introduction}
The availability of large amounts of labeled data remains a key factor in the successful training of deep neural network classification models using supervised learning. 
In contrast, the human brain has the remarkable ability to learn new classes with just a few (visual) examples, or even with no examples at all but with
the use of other data modalities such as verbal descriptions of these  new classes. 
This paper is concerned with the \emph{zero-shot learning} (ZSL) paradigm whose goal is to achieve fast learning, where
the learner is expected to recognize and correctly predict classes not present in the training data.

%


To tackle the challenging task of ZSL, we first need to establish a reference space in which any potential class of interest can be defined. The current literature offers two approaches to building this reference space, which is often referred to as the \emph{semantic space}.
The first approach is 
to encode the description of each class with a language model \cite{reed2016learning,xian2017zero}. This  approach assumes that the language model is sufficiently expressive and can associate the attributes of a class with the class name itself, i.e., the language model builds an embedding space where the embedding of a class name also encodes the attributes of the class, and clusters \emph{visually} similar concepts together. 
The second approach to defining class is  via a set of attributes, and build binary or continuous vectors that define the classes (a.k.a. \emph{class attribute vectors}). All standardized ZSL benchmark datasets (CUB \cite{WahCUB_200_2011}, AWA2 \cite{xian2017zero} and SUN \cite{patterson2012sun}) come with such class defining vectors that are constructed using a set of predefined attributes. Most contemporary approaches use these supplied attribute vectors.

Published ZSL solutions can be divided into two groups. Solution in the first group are \emph{generative or hallucinative} methods \cite{xian2019f,verma2018generalized,schonfeld2019generalized,narayan2020latent}, 
which try to learn a generator that generates examples conditioned on the class defining vectors. Once the generator has converged, it is used to generate examples of seen and unseen classes. After acquiring the generated samples, the above mentioned approaches fall back to the fully supervised setting, where they learn a fully supervised classifier using the generated data. These approaches depend heavily on the quality of the generated samples and the generator's ability to approximate the classes' real data feature distributions. Such approaches suffer from the shortcomings of whichever GANs and VAEs are used, e.g., mode collapse, and the difficulty in training and converging.
The second group include \emph{discriminative} approaches \cite{Huynh_2020_CVPR,xu2020attribute,liu2021goal} which tries to learn a compatibility function that measures the compatibility of a sample's embedding with a class definition vector (typically an inner product or a nearest-neighbor are used to estimate compatibility), and classify an input sample as the class with the highest compatibility score. The objective in these frameworks is to learn a good mapping from the visual space (where  image features reside) to the semantic space (where class definitions reside). They usually suffer from a substantial bias  towards the seen classes, which can be alleviated using a calibrated stacking ``trick'' \cite{chao2016empirical,xu2020attribute}, or a customized loss term \cite{Huynh_2020_CVPR}.

This paper approaches ZSL through the discriminative framework and introduces a novel mechanism aiming at improving the mapping between the visual and semantic spaces. We do so with sample augmentation in the deep visual embedding space via dropout and map the new augmented sample to a new fictitious or imagined class in the semantic space that corresponds to it. This method exposes the learning system to more classes and more combinations of attributes than it originally has from the training set.

To summarize, our main contribution is
the design of a new mechanism that can potentially help any 
discriminative zero-shot learning model to generalize better to unseen classes by exposing the model to new fictitious classes during training. We test the proposed mechanism over the DAZLE method 
of \cite{Huynh_2020_CVPR}, which leads to improving the state-of-the-art ZSL performance on the CUB dataset, and improving the performance of the DAZLE model itself on all standardized ZSL benchmark datasets. 
Finally, we point out that catastrophic forgetting is 
one of the obstacles that end-to-end zero shot learners should consider and observe that its severity depends the backbone architecture and the dataset used.


\section{Related Work}
\label{Related Work}
As mentioned in the introduction, ZSL methods can be broadly divided into two families: generative and discriminative models.
Representatives of the generative approach include: \cite{xian2019f, verma2018generalized, schonfeld2019generalized, narayan2020latent}. In \cite{verma2018generalized}, the authors link a CVAE and a regressor network. The CVAE encodes and reconstructs the input features conditioned on the input's true class definition vector, and the regressor tries to predict the attribute vector from the generated sample. In \cite{xian2019f}, the authors link a CVAE and a GAN by using the same generator for both. The CVAE tries to reconstruct the input feature vector, conditioned on the class definition vector of the sample, and the generated or reconstructed sample is passed to a discriminator that tries to distinguish true samples from generated ones. The authors also add a second discriminator for the transductive ZSL setting that tries to discriminate between generated samples for unseen samples and real unlabeled samples of unseen classes. In \cite{schonfeld2019generalized}, the authors propose to learn two VAEs---one that learns to reconstruct the visual space and  another that learns to reconstruct the semantic space. These VAEs are linked together by cross alignment, meaning the encoded latent random vector from the visual VAE should be able to reconstruct the class vector in the semantic space and vice versa, and distribution alignment, which penalizes the model with the distance between the learned visual features distribution and the semantic features distribution. \cite{narayan2020latent} propose an architecture similar to \cite{xian2019f} but add a feedback module that tries to correct the intermediate representation in the generator using information from the discriminator.

Papers \cite{Huynh_2020_CVPR,xu2020attribute,liu2021goal} are representatives of the second group.
Their models use the class definition vectors as a fixed classification layer, and add modules that help the network  localize and implicitly detect attributes in the visual space. We focus on and extend \cite{Huynh_2020_CVPR} in our Method section, since it does not require extra knowledge such as human gaze points as leveraged by \cite{liu2021goal}, or too many added loss terms to fine-tune as proposed by \cite{xu2020attribute}.

\section{Problem setting}
\label{Problem Setting}

In this section, we describe the problem definition, notations used throughout the paper, and evaluation metrics that are widely used in the ZSL literature.
Let $\scriptC$  be the set of classes in our world (e.g., $\{dogs,cats,okapi, \ldots \}\subseteq \scriptC$). In the ZSL problem, the learning system is presented with a limited number of seen classes, $\scriptC_{seen}=\{c_i \}_{i=1}^{N_{s}}\subset \scriptC$, through the training dataset $\scriptD_{tr}=\{(x_i,y_i) |  y_i\in \scriptC_{s} \}_{i=1}^{N_{tr}}$, and is expected to identify a wider range of classes $\scriptC= \scriptC_{s}\cup \scriptC_{u}$. In classical zero-shot learning (CZSL) we only care about the learner's ability to recognize the unseen classes, while in the generalized zero-shot problem the learner is expected to perform well on both seen and unseen classes. 

The GZSL problem is considered a more realistic problem since we are more likely to encounter both seen and unseen classes in real world scenarios. It is also considered the harder variant because of the inherent bias toward seen classes that most methods have.

The knowledge transfer from seen and unseen classes is usually done  through a semantic space, so in addition to the training dataset $\scriptD_{tr}$, the learner also receives class definitions. These class definitions, as mentioned earlier, can either be a simple class name embedding through an NLP model, or handcrafted vectors that define classes in the world via a set of $n$ predefined attributes. In either case, we denote the class definition vector as $a=\phi(y) \in \Reals^n$.

\subsection{Evaluation metrics}
\label{evaluation metrics}

In the GZSL literature, three metrics are widely used to evaluate the proposed methods. These are the per-class accuracy on seen classes $acc_s$, the per-class accuracy on unseen classes $acc_u$ and the harmonic mean of these two accuracies $Hm$:

$$
Hm(acc_u, acc_s) = \frac{2 acc_u \times acc_s}{(acc_u + acc_s)}.
$$ 

The per-class accuracy of a group of classes is defined as the average accuracy of each of the member classes. For  example, per-class accuracy on unseen classes is defined as,
$$
acc_u = \frac{1}{|\scriptC_u|}\sum_{c\in \scriptC_u} \frac{\text{\#samples correctly predicted in } c}{\text{\#samples in } c} .
$$

Note that the prediction space in the GZSL problem is $\scriptC_{s}\cup \scriptC_{u}$, which means that the learner can mispredict a test sample from the seen classes as being one of the unseen classes and vice versa.

The per-class accuracy is chosen instead of the regular per-sample accuracy in order to eliminate the effect of class size on the final accuracy. We want the learner to preform well on all classes no matter the relative size of the class in the test set.
A similar rationale justifies the usage of the harmonic mean instead of the regular arithmetic mean, since we want the learner to preform well on both groups of classes ($\seenC$ and $\unseenC$).

Another metric that is widely reported in the literature is \emph{zero-shot accuracy} for CZSL. 
Denoted $\text{T}1$, this metric is the per-class accuracy on unseen classes, \emph{but} the prediction space is only $\unseenC$, and the test set only contains unseen class samples.

\section{Using fictitious classes}
\label{Method}
Our method can be used to extend any discriminative ZSL model that has implicit attribute detection and uses class definition vectors as a fixed classification layer (e.g., \cite{Huynh_2020_CVPR,xu2020attribute,liu2021goal}). 
Before explaining our method, we start by describing the DAZLE model \cite{Huynh_2020_CVPR}, Adding fictitious classes, easy-to-understand and has publicly available implementation. After that we describe how to apply our method on top of DAZLE.

\subsection{DAZLE}
\label{DAZLE}

\begin{figure}
  \centering
  \includegraphics[width=\linewidth]{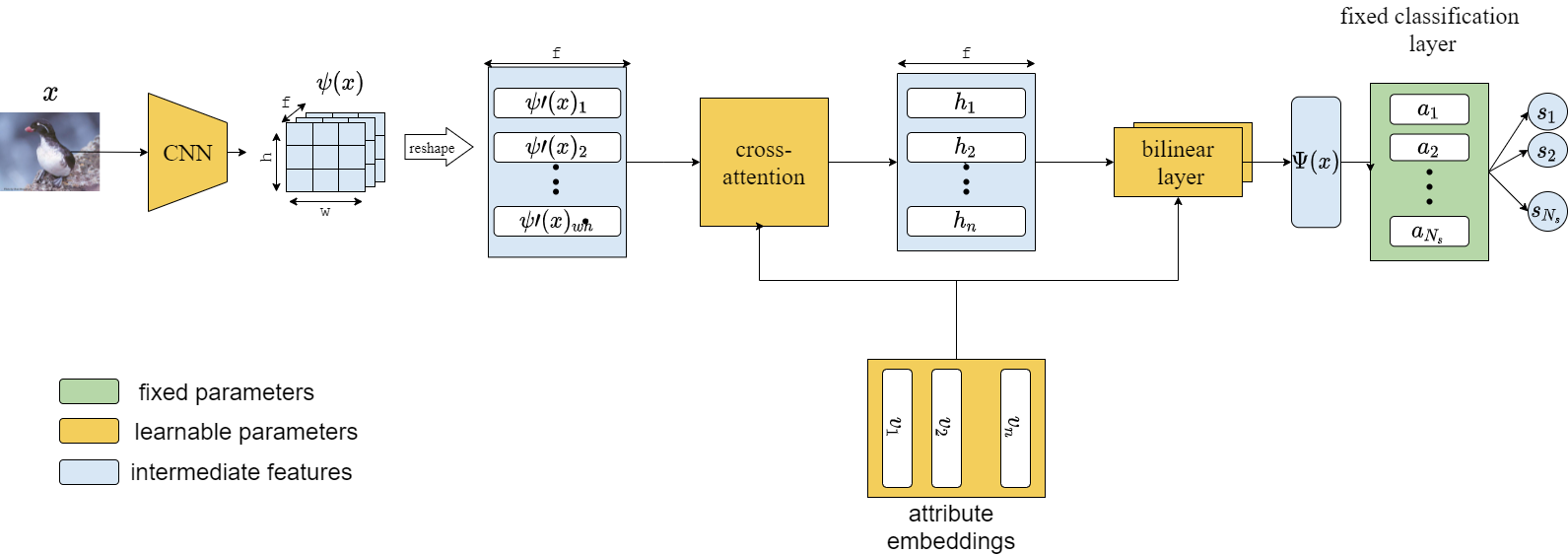}
  \caption{Our illustration of the DAZLE model \cite{Huynh_2020_CVPR}}
  \label{fig:dazle_figure}
\end{figure}

DAZLE \cite{Huynh_2020_CVPR} is a novel and intuitive architecture to train and solve GZSL, which produced state-of-the-art results as Aug. of 2020 against standard GZSL benchmarks. The architecture is shown in Figure \ref{fig:dazle_figure}.
The input of the proposed architecture is the conv-map features of an image $x$ extracted from the last convolutional layer of an ImageNet pretrained network. We denote the conv-map by $\psi(x)\in \Reals^{w\times h\times f}$). Choosing the output of a convolutional layer as the input
allows the learning system to retain spatial information and attend differently to different regions in the image and associate different attributes to them. After extraction from the backbone, the conv-map is reshaped to be a $2d$ matrix of $wh$ rows that embed the $wh$ image regions in $f$ columns (denoting the reshaped version as $\psi^\prime(x)\in \Reals^{wh\times f}$). Next, a cross-attention mechanism is applied between the feature vectors of the regions and learned embedding vectors that represent the attributes or the building blocks of classes in the world. The equations for DAZLE's attention mechanism are:
\begin{eqnarray*}
A_{i,j} &=& \frac{e^{\psi^\prime(x)_i^T W_{\alpha} v_j}}{\Sigma_r{e^{\psi^\prime(x)_r W_{\alpha} v_j}}} \\
h_{j} &=& \Sigma_r{A_{r,j}\psi^\prime(x)_r} ,
\end{eqnarray*}
where $A_{i,j}$ is the attention score between region $i$, represented by $\psi^\prime(x)_i$, and the attribute $j$ represented by $v_j$, $W_{\alpha}$ is the learned attention weight matrix, and $h_{j}$ is the output of the cross-attention block, which essentially represents the attribute $j$ in the visual space as a weighted sum of the regions' feature vectors.
After the attention block, the attributes visual representatives $h_{j}$ are passed in parallel through two parallel bi-linear layers with weights $W_{e}, W_{\beta}$, where one of them is gated with a sigmoid. 
The outputs $z, e\in \Reals^n$  of these two layers are
\begin{eqnarray*}
z_j &=& v_j^T W_\beta h_j \\
e_j &=& \sigma(v_j^T W_e h_j)
\end{eqnarray*}
Cell $j$ of these vectors represents the score or existence of attribute $j$ in the given image $x$.
Finally, we obtain the embedding of the image $x$ in terms of our $n$ attributes as the element-wise multiplication of $z$ and $e$,
$$
\Psi(x) = z\odot e \in \Reals^n .
$$
To get the classification scores for image $x$, we use the class definition vectors as classifiers, so the classification score for image $x$ as class $k$ is
$ s_k = \Psi(x)^T\phi(y_k)$.

The authors of \cite{Huynh_2020_CVPR} also describe a way of augmenting the classification scores with a fixed bias and a new loss term $\script[L]_{cal}$ that tries to reduce the inherent bias toward seen classes. We do not use this in our method since it requires knowledge about the targeted unseen classes during training. Instead, we use the calibrated stacking to re-balance the model. The calibrated stacking trick augments the classes scores $\{s_i\}$ by subtracting a bias constant $\gamma$ from the seen classes scores of the given instance:
$$
\Tilde{s}_i = s_i - \gamma\mathbbm{1}_{i\in\seenC} .
$$
Thus re-calibrating the model and reducing its bias to the seen classes (the model will be less likely to mispredict unseen class instance as a seen class).
A detailed description of our implementation of DAZLE appears in Section~\ref{changes_to_DAZLE}.
We use our version of DAZLE in all our experiments.

\subsection{Dropout and fictitious classes}
\label{dropout and fake classes}

\begin{figure}
\centering
\begin{subfigure}{.34\textwidth}
  \centering
  \includegraphics[width=\linewidth]{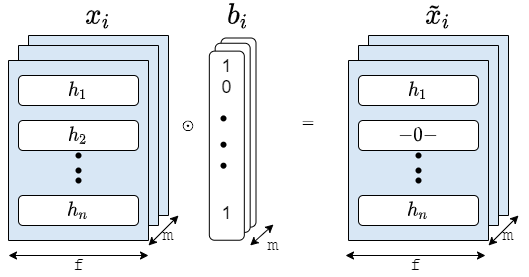}
  \caption{Creating augmented samples to be added to the batch}
  \label{fig:method_create_samples}
\end{subfigure}%
\hspace{.05\linewidth}
\begin{subfigure}{.3\textwidth}
  \centering
  \includegraphics[width=\linewidth]{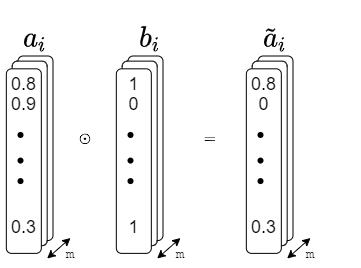}
  \caption{Creating corresponding\\ fictitious classes}
  \label{fig:method_create_classes}
\end{subfigure}%
\begin{subfigure}{.3\textwidth}
  \centering
  \includegraphics[width=0.5\linewidth]{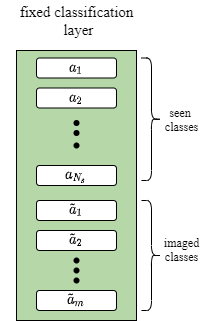}
  \caption{Adding the generated classes to the fixed classification layer}
  \label{fig:method_add_classes}
\end{subfigure}
\caption{Creation and usage of fictitious classes: we use the same dropout mask that augmented teh samples to create the corresponding fictitious class definitions which we add them to the fixed classification layer}
\label{fig:our_method}
\end{figure}

Our method is based on a simple idea. Similar to other augmentation-based regularizers, we want to expand the number of available training examples. Nevertheless, unlike most other augmentation methods that assign the augmented version of the sample to the same class, we assign it to a new class that is an augmented version of the original sample's class. Thus, theoretically we can explore a much larger group of classes during training, and are not confined to the set of base seen classes.

From the training batch we sample $m$ samples with replacement to augment, meaning a sample could be chosen twice to be augmented. The augmentation is done on the output of the attention block in DAZLE (Figure \ref{fig:method_create_samples}). Looking at the output of the attention mechanism in DAZLE, we have a matrix $H = \text{concat}(\{h_j\}_{j=1}^n) \in \Reals^{n\times f}$ per sample, where each row is a vector representing an attribute in the visual space. We sample a binary vector $b \sim Ber(p)^n$ per chosen sample that decides which rows to drop from $H$, then add the augmented samples back to the batch, which continues to pass through the remaining layers. 
Now, in contrast to the normal dropout, we save the the binary masks in order to augment the classifiers and add the augmented classifiers to the rest of base seen classifiers  (Figure \ref{fig:method_create_classes}). More precisely, augmenting $m$ samples would result in adding $m$ new classifiers to the fixed classification layer. To obtain the augmented classifier that corresponds to an augmented sample $\Tilde{x_i}$, we take the class definition of the original sample $a_i=\phi(y_i)$ and augment it with the same binary $b_i$ mask that generated $\Tilde{x}_i$ from $x_i$, (i.e., $\Tilde{a}_i = b_i\odot a_i$). This makes sense because if we dropped the visual representative of an attribute, the sample would not belong to the original class; it would belong to a similar class that does not have the missing attribute. Thus, we should also drop the corresponding cell of the attribute in the class definition vector.

Note that for each given training batch, the classification layer is changed to contain $N_s+m$ classifiers, $N_s$ classifiers for the original seen classes, and $m$ classifiers for the new generated classes. Consequently, the prediction space for a training batch is the group of seen and generated classes (Figure~\ref{fig:method_add_classes}). 
Both $m$ (the number of new fake classes per batch), and $p$ (the dropout rate) are two hyperparameters of our method and we later 
investigate their effect.

\subsection{Training}
\label{training}
We use categorical cross-entropy to train the model. We first freeze the CNN backbone and train the DAZLE modules until they converge; then we unfreeze the CNN backbone and continue training using a lower learning rate with a gradient flowing to all layers. This training strategy is a double-edged sword: On one hand, it allows us to fine-tune the backbone, which updates the CNN kernels to be more suited to the new domain and achieve higher accuracies on seen classes, but on the other hand, the CNN may forget important patterns that are essential to recognize unseen classes. This phenomenon has long been known in the lifelong learning literature and is dubbed  Catastrophic Forgetting. We expand on catastrophic forgetting in Section \ref{Experiments}.
We train with frozen backbone for 30 epochs and then unfreeze it and continue training for another 50 epochs. The rest of the hyper-parameters: $m,p$ of our method as well as $\gamma$ for calibrated stacking for each dataset were chosen such that they maximize the harmonic mean over the 3 folds from the validation sets as defined by \cite{xian2017zero}. a table containing those hyperparameters can be found in the supplementary material \ref{hyperparameters}.

\section{Experiments}
\label{Experiments}

\subsection{Datasets}
\label{Datasets}
The three datasets that are used as benchmarks in the ZSL literature are CUB \cite{WahCUB_200_2011}, AWA2 \cite{xian2017zero} and SUN \cite{patterson2012sun}. CUB is a fine-grained dataset, meaning the dataset contains very visually similar classes, it containing 200 bird classes with 312 attributes. 150 of out of the 200 classes are considered seen classes that we are free to train on, while the remaining 50 classes are unseen classes and should not be trained on. AWA2 is a coarse-grained (e.g., most classes aren't visually similar or related), unbalanced dataset of animals with 85 attributes that is split into 40 seen classes and 10 unseen classes. SUN is a scene understanding dataset with 102 attributes that has 645 seen classes and 72 unseen classes. The splits between seen and unseen classes were standardized in \cite{xian2017zero} and chosen such that the unseen classes group in each dataset does not intersect with the 1000 Image-Net classes that most CNN backbones are pre-trained on and used for feature extraction. Thus, the unseen classes are truly unseen to the whole learning pipeline. including the pre-trained part.

\begin{table}
  \caption{Comparison of various ZSL methods. Blue numbers correspond to best performance and red correspond to second.
  The first section contains discriminative methods, the second section contains generative methods and the third one contains end-to-end trained methods. Baseline results are taken from the corresponding papers (missing performance numbers were not reported).}
  \label{table_comparison}
  \centering
  \resizebox{\linewidth}{!}{%
  \begin{tabular}{lllllllllllll}
    \toprule
    \multirow{2}{4em}{Methods} & \multicolumn{4}{c}{CUB}    &  \multicolumn{4}{c}{SUN}  &      \multicolumn{4}{c}{AWA2}\\
      \cmidrule(r){2-5}  \cmidrule(r){6-9}  \cmidrule(r){10-13}
      &T1     &  $acc_u$     & $acc_s$  & Hm & T1     &  $acc_u$     & $acc_s$  & Hm & T1     &  $acc_u$     & $acc_s$  & Hm\\
    
    \midrule
PSR\cite{annadani2018preserving}    &56.0    &24.6    &54.3    &33.9    &61.4    &20.8    &37.2    &26.7    &63.8    &20.7    &73.8    &32.3\\
RN\cite{sung2018learning}           &55.6    &38.1    &61.1    &47.0    &-    &-    &-    &-    &64.2    &30.0    &\colorblue[93.4]    &45.3\\
SP-AEN\cite{chen2018zero}           &55.4    &34.7    &70.6    &46.6    &59.2    &24.9    &38.6    &30.3    &-    &-    &-    &-\\
IIR\cite{cacheux2019modeling}       &63.8    &55.8    &52.3    &53.0    &63.5    &47.9    &30.4    &36.8    &67.9    &48.5    &83.2    &61.3\\
TCN\cite{jiang2019transferable}     &59.5    &52.6    &52.0    &52.3    &61.5    &31.2    &37.3    &34.0    &71.2    &61.2    &65.8    &63.4\\
E-PGN\cite{yu2020episode}           &72.4    &52.0    &61.1    &56.2    &-    &-    &-    &-    &73.4    &52.6    &83.5    &64.6\\
DAZLE\cite{Huynh_2020_CVPR}             &65.9    &56.7    &59.6    &58.1    &-    &52.3    &24.3    &33.2    &-    &60.3    &75.7    &67.1\\
\midrule
f-CLSWGAN\cite{xian2018feature}         &57.3    &43.7    &57.7    &49.7    &60.8    &42.6    &36.6    &39.4    &-    &-    &-    &-\\
cycle-CLSWGAN\cite{felix2018multi}      &58.4    &45.7    &61.0    &52.3    &60.0    &49.4    &33.6    &40.0    &-    &-    &-    &-\\
CADA-VAE\cite{schonfeld2019generalized} &-       &51.6    &53.5    &52.4    &-       &47.2     &35.7    &40.6    &-    &55.8    &75.0    &63.9\\
OCD-CVAE\cite{keshari2020generalized}   &60.3    &44.8    &59.9    &51.3    &{\colorred[63.5]}    &44.8    &\colorred[42.9]    &43.8    &71.3    &59.5    &73.4    &65.7\\
RFF-GZSL(1-NN)\cite{han2020learning}    &-       &50.6    &79.1    &61.7       &-    &{\colorblue[56.6]}    &42.8    &\colorred[48.7]    &-    &-    &-    &-\\
IZF\cite{shen2020invertible}         	&67.1    &52.7    &68.0    &59.4    &{\colorblue[68.4]}    &{\colorred[52.7]}    &\colorblue[57.0]    &\colorblue[54.8]    &\colorblue[74.5]    &60.6    &77.5    &68.0\\
LsrGAN\cite{vyas2020leveraging}     &60.3    &48.1    &59.1    &53.0    &62.5    &44.8    &37.7    &40.9    &-    &-    &-    &-\\
\midrule
QFSL\cite{song2018transductive}     &58.8    &33.3    &48.1    &39.4    &56.2    &30.9    &18.5    &23.1    &63.5    &52.1    &72.8    &60.7\\
LDF\cite{li2018discriminative}      &67.5    &26.4    &{\colorblue[81.6]}    &39.9    &-    &-    &-    &-    &-    &-    &-    &-\\
SGMA\cite{zhu2019semantic}          &71.0    &36.7    &71.3    &48.5    &-    &-    &-    &-    &-    &-    &-    &-\\
AREN\cite{xie2019attentive}         &71.8    &63.2    &69.0    &66.0    &60.6    &40.3    &32.3    &35.9    &67.9    &54.7    &79.1    &64.7\\
LFGAA\cite{liu2019attribute}        &67.6    &36.2    &{\colorred[80.9]}    &50.0    &61.5    &18.5    &40.0    &25.3    &68.1    &27.0    &\colorblue[93.4]    &41.9\\
DVBE\cite{min2020domain}            &-       &64.4    &73.2    &68.5    &-       &44.1    &41.6    &42.8    &-       &62.7    &77.5    &69.4\\
RGEN\cite{xie2020region}            &76.1    &60.0    &73.5    &66.1    &63.8    &44.0    &31.7    &36.8    &\colorred[73.6]    &\colorblue[67.1]    &76.5    &\colorblue[71.5]\\
APN \cite{xu2020attribute}         	&72.0    &65.3    &69.3    &67.2    &61.6    &41.9    &34.0    &37.6    &68.4    &56.5    &78.0    &65.5\\
GEM-ZSL \cite{liu2021goal}     	    &77.8    &64.8    &77.1    &70.4    &62.8    &38.1    &35.7    &36.9    &67.3    &\colorred[64.8]    &77.5    &\colorred[70.6]\\
\midrule
\midrule

Our method (DT3)       &{\colorblue[80.1]}    &{\colorred[72.4]}   &75.6   &{\colorblue[74.0]}       &58.8   &40.3   &32.1   &35.7      &68.5	&59.2	&80.4	&68.2 \\
Our method (RL3)       &{\colorred[79.9]}     &{\colorblue[72.9]}   &73.1   &{\colorred[73.0]}       &60.9   &42.5   &31.2   &36.0      &62.8   &56.5   &76.6   &65.0 \\
Our method (RL4)       &76.1     &67.8   &74.3   &70.9       &62.9   &47.8   &36.0   &41.1      &63.5   &53.0   &\colorred[86.2]   &65.6 \\

    \bottomrule
  \end{tabular}%
  } 
\end{table}%

\subsection{Comparison to state of the art}
\label{comparison to state of the art}
In Table \ref{table_comparison} we see the results the of most recent ZSL methods. The first section contains discriminative methods, the second section contains generative methods and the third contains end-to-end trained methods. 
Table~\ref{table_comparison} is taken from \cite{liu2021goal} and extended using our results. We add to the table the results of our method when applied to DAZLE. We report results when using different backbones to the model. Using DenseNet 201 3rd transition layer (denoted DT3) as the backbone gives the best results with our method in two out of the 3 datasets (CUB and AWA2). While using ResNet 101 the 4th and final  convolution block (denoted RL4) gives the best result on SUN. For each metric we highlight in blue the best result and in red the second best result.


The proposed method surpasses previous state-of-the-art performance on CUB with 74.0 harmonic mean (Hm) accuracy in the GZSL setting, and 80.1 zero-shot accuracy (T1), compared to the previous best result of 70.4 in Hm and 77.8  in T1 \cite{liu2021goal}.
In AWA2, our method is only comparable to the competition's results but does not surpass it.

For the SUN dataset we could not improve the state-of-the-art, but 
 our method is comparable to other discriminative and end-to-end ZSL methods (such as APN\cite{xu2020attribute} and GEM-ZSL\cite{liu2021goal}). The state-of-the-art performance 
 in the SUN dataset has been achieved by the flow-based model IZF
 of Shen et al. \cite{shen2020invertible}.

\subsection{Catastrophic forgetting}
\label{Catastrophic forgetting}

\begin{figure}
\centering
\begin{subfigure}{.35\textwidth}
  \centering
  \includegraphics[width=\linewidth]{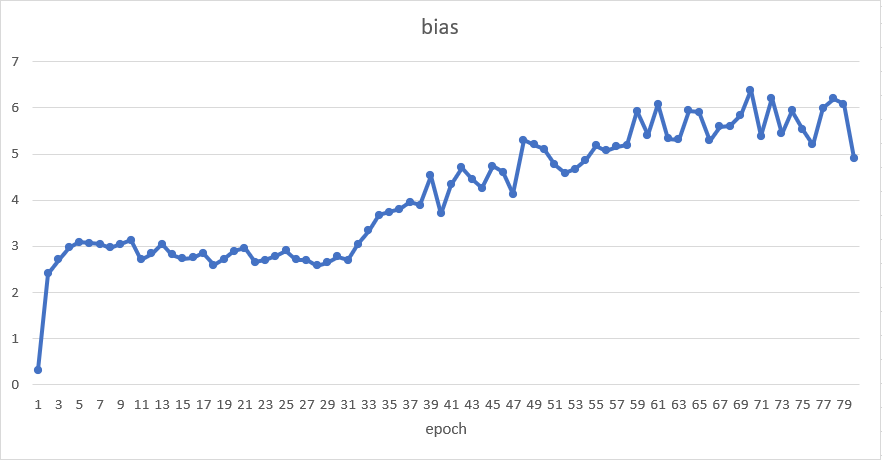}
  \caption{$\gamma$-bias toward seen classes}
  \label{fig:cf bias}
\end{subfigure}%
\begin{subfigure}{.35\textwidth}
  \centering
  \includegraphics[width=\linewidth]{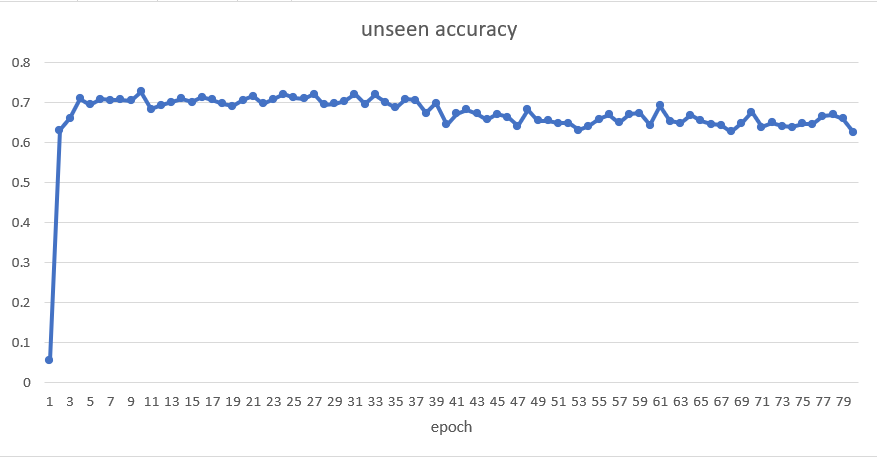}
  \caption{Unseen accuracy}
  \label{fig:cf unseen}
\end{subfigure}%
\begin{subfigure}{.35\textwidth}
  \centering
  \includegraphics[width=\linewidth]{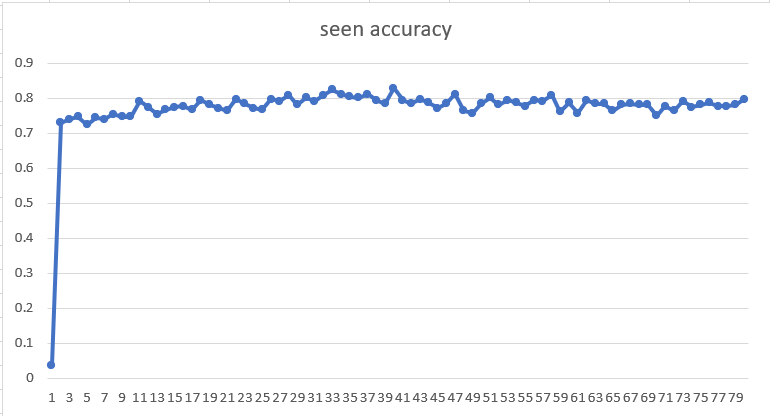}
  \caption{Seen accuracy}
  \label{fig:cf seen}
\end{subfigure}
\caption{Catastrophic forgetting as observed on the AWA2 validation set using the ResNet backbone. We see that the bias toward seen classes increases after allowing the gradient to propagate through the backbone, and the how the accuracy on unseen classes $acc_u$ starts to drop as the training continues but the accuracy on seen classes $acc_s$ increases and plateaus.}
\label{fig:catastrophic_forgetting_awa2}
\end{figure}

\begin{figure}
\centering
\begin{subfigure}{.35\textwidth}
  \centering
  \includegraphics[width=\linewidth]{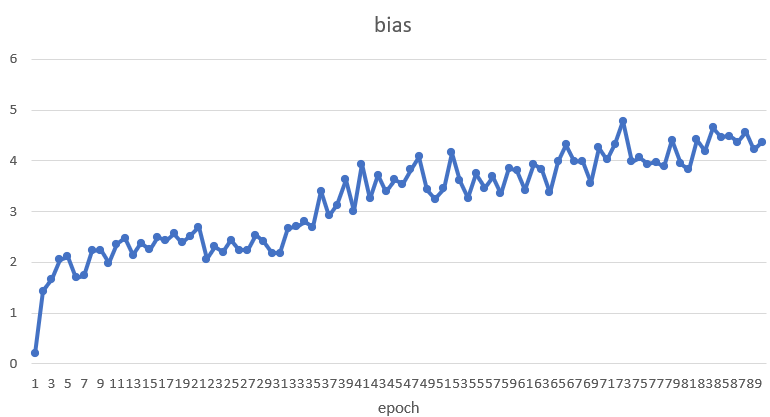}
  \caption{$\gamma$-bias toward seen classes}
  \label{fig:cf bias 2}
\end{subfigure}%
\begin{subfigure}{.35\textwidth}
  \centering
  \includegraphics[width=\linewidth]{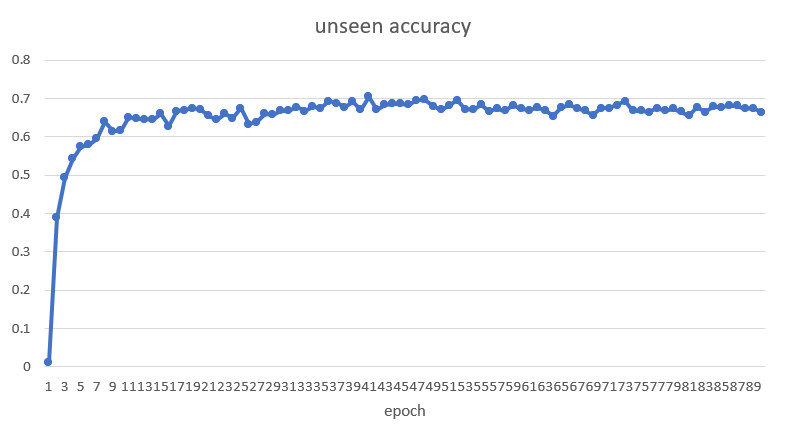}
  \caption{Unseen accuracy}
  \label{fig:cf unseen 2}
\end{subfigure}%
\begin{subfigure}{.35\textwidth}
  \centering
  \includegraphics[width=\linewidth]{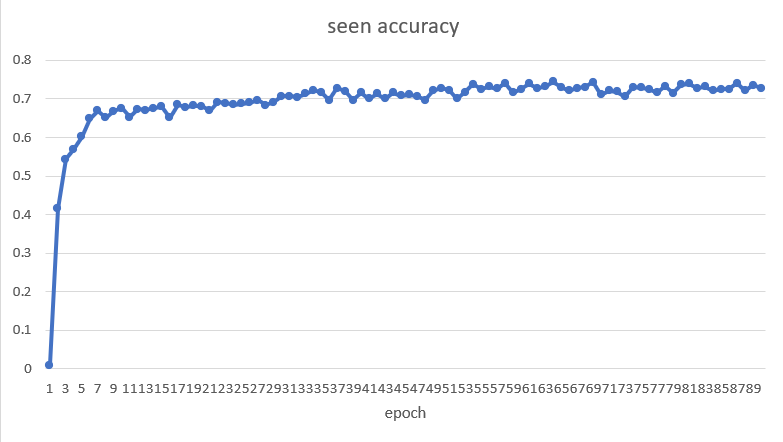}
  \caption{Seen accuracy}
  \label{fig:cf seen 2}
\end{subfigure}
\caption{Catastrophic forgetting as observed on the CUB validation set using the DenseNet backbone.
The bias toward seen classes still increases as the gradients start to update the backbone, but  the performance in $acc_u$  does not drop as sharply.}
\label{fig:catastrophic_forgetting_cub}
\end{figure}

Catastrophic forgetting was first observed in \cite{french1999catastrophic}. And is usually mentioned in lifelong  learning literature (\cite{goodfellow2013empirical, kemker2017fearnet, kemker2018measuring, kirkpatrick2017overcoming}) where it is described as the inability to retain performance on previously learned tasks after training on a new task. In \cite{kemker2018measuring}, the authors explain that catastrophic forgetting occurs when training on the new task causes large weight changes. Those large Weight changes, in return,  cause disruptions to representations learned for previous tasks, which leads to performance degradation on the previously learned task. 

In our context, the first learned task would be ImageNet (since we use pretrained CNNs), and the second task would be the seen classes of the target dataset. But we define catastrophic forgetting a bit differently. We define catastrophic forgetting as the inability to retain performance on unseen classes after backbone fine-tuning while keeping or improving performance on seen classes. We hypothesize that this phenomenon happens for the same reason that causes catastrophic forgetting in the lifelong learning context. Fine-tune the CNN backbone causes it to forget important kernels or patterns that are crucial in distinguishing unseen classes. What this means is that, counter-intuitively, a vanilla pre-trained CNN backbone may perform better on the unseen classes than a fine-tuned one. To the best of our knowledge, we are the first to observe and point out this problem when training an end-to-end deep zero-shot model.


As mentioned before (in Section \ref{training}), we experienced catastrophic forgetting with our model  as well. 
We notice this phenomenon in various degrees depending on the dataset and the backbone architecture chosen. Nevertheless, generally speaking, DenseNet is more resilient to catastrophic forgetting than ResNet. and AWA2 is the dataset that suffers the most from it.
In Figure \ref{fig:catastrophic_forgetting_awa2}, we see how the accuracy on the unseen classes drops by ~10\% after allowing the gradient to change the backbone weights, while the performance on seen classes improves and plateaus. We also notice how the optimal bias constant $\gamma$ starts increasing after allowing the gradient to flow to the CNN (after epoch 30), which means that the inherent bias toward seen classes of the model increases as well.
As a comparison, in Figure~\ref{fig:catastrophic_forgetting_cub}, we show that catastrophic forgetting is not as severe on the CUB validation set with the DenseNet backbone. Even when the performance on unseen classes does not drop as sharply, we still notice that the bias toward seen classes increases after allowing the gradient to flow back to the backbone.
We note that the increased bias toward seen classes makes it harder to chose the bias correction constant $\gamma$ from the validation set to apply it to the test set.  And in that regard, we recommend choosing multiple $\gamma$ constants from the validation set's bias curve use them incrementally as the training progresses.

\subsection{Toy example}
\label{Toy example}

\begin{figure}[htb]
\centering
\begin{subfigure}{.5\textwidth}
  \centering
  \includegraphics[width=\linewidth]{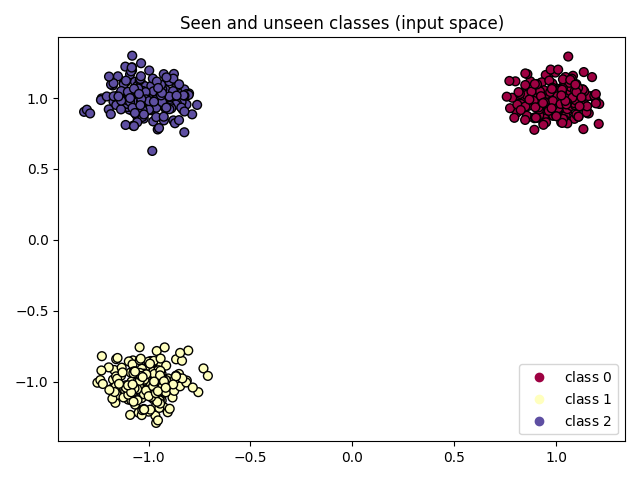}
  \caption{Seen and unseen classes.}
  \label{fig:Seen and unseen classes}
\end{subfigure}%
\begin{subfigure}{.5\textwidth}
  \centering
  \includegraphics[width=\linewidth]{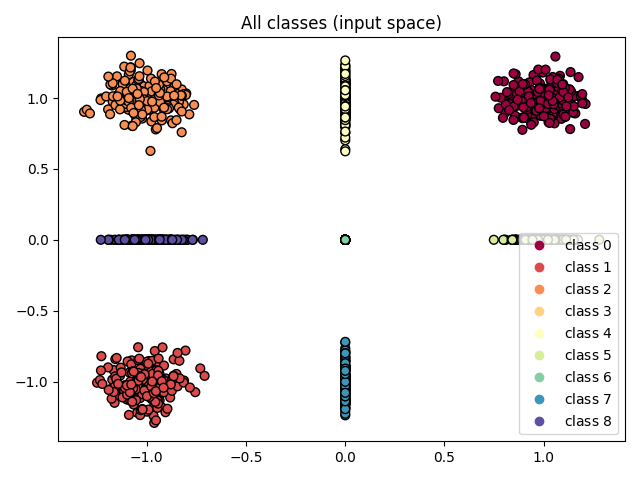}
  \caption{All classes.}
  \label{fig:All classes}
\end{subfigure}
\caption{Toy dataset we use to illustrate the effect of our method: the two seen classes generated from an isotropic Gaussian with 0.1 variance centered at $[(1,1), (-1,-1)]$ and the seen class is generated from third Gaussian with the same variance centered at $(-1,1)$. The fictitious classes samples are generated from applying dropout to the seen classes samples.
(best viewed in color).}
\label{fig:toy_example_dataset}
\end{figure}

To visualize and demonstrate our method, we apply it to a toy dataset. The dataset can be seen in Figure \ref{fig:Seen and unseen classes}. The dataset is generated by three two-dimensional isotropic Gaussians with 0.1 variance centered at points $([1,1],[-1,-1],[-1,1])$. The first two will be considered seen classes and the third is the unseen class. Applying our method will result in adding five more fictitious classes, which can be seen in Figure \ref{fig:All classes}. 

We train three models on this toy dataset and use calibrated stacking to balance them. The first model is the vanilla model that only trains on the seen classes. The second model uses our method to augment the seen classes and trains on the set of seen and fictitious classes. In addition to these two models, we also train a third model that is almost like the first one. It only sees the seen classes but uses the regular dropout to augment the samples and maps them to the original class. 

\begin{figure}
\centering
\begin{subfigure}{.35\textwidth}
  \centering
  \includegraphics[width=\linewidth]{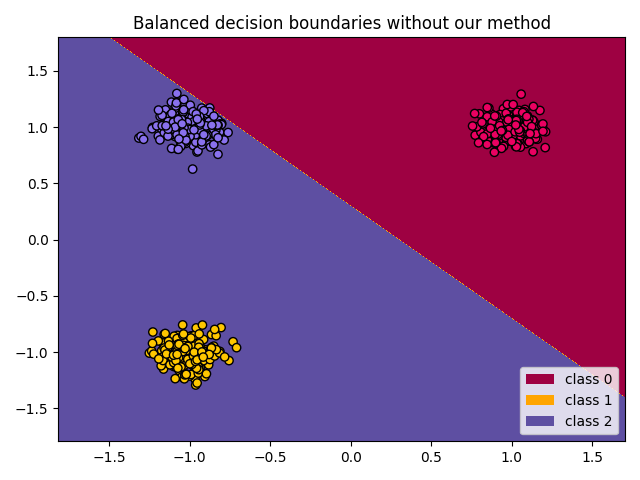}
  \caption{Vanilla.}
  \label{fig:db vanilla}
\end{subfigure}%
\begin{subfigure}{.35\textwidth}
  \centering
  \includegraphics[width=\linewidth]{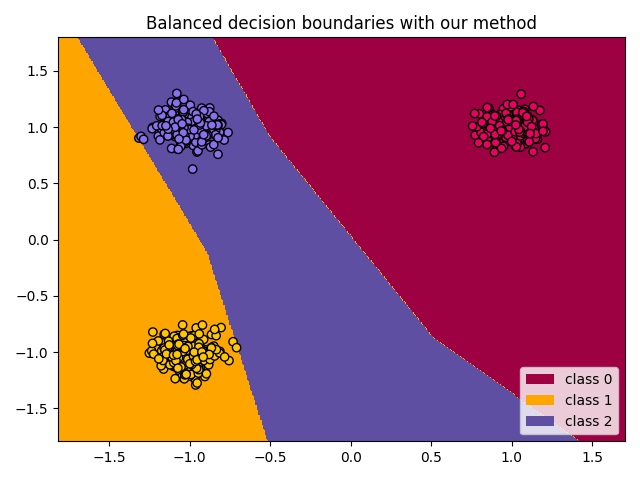}
  \caption{Our method.}
  \label{fig:db ours}
\end{subfigure}%
\begin{subfigure}{.35\textwidth}
  \centering
  \includegraphics[width=\linewidth]{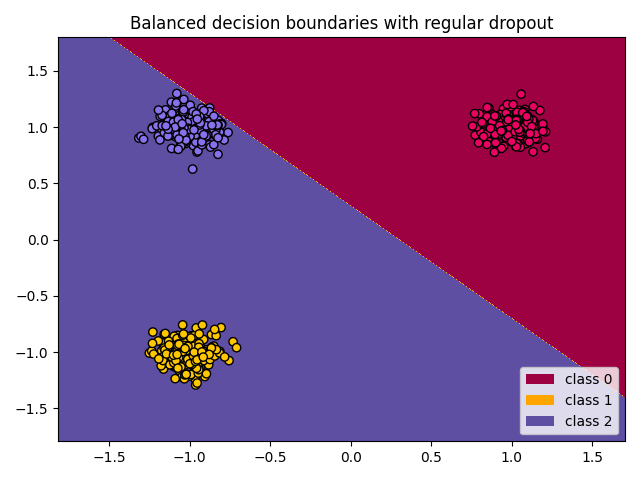}
  \caption{Regular dropout.}
  \label{fig:db reg dropout}
\end{subfigure}
\caption{Decision boundaries we get when training with different methods on the toy dataset and using calibrated stacking. (best viewed in color).}
\label{fig:toy_example_boundries}
\end{figure}

The decision boundaries resulting from each of the three models are shown in Figure \ref{fig:toy_example_boundries}. As we can see, in both the vanilla and regular dropout models, the learned decision boundary is linear, which makes sense because these models only saw the seen classes that could be separated by a linear boundary using a single feature (distance from the $y=-x$ plane). Our model, on the other hand, learns more complex decision boundaries since it saw more classes during training and learned that the distinguishing features of the seen classes are not just the distance from a linear plane.

\subsection{Ablation study}
\label{ablation_study}

\begin{table}[]
\caption{The effects of our method's components measured on all three datasets with a 3-fold validation sets average.}
\label{table_ablation}
\begin{center}
\resizebox{\linewidth}{!}{%
\begin{tabular}{@{}lllllll@{}}
 \toprule
\multicolumn{1}{c}{\multirow{2}{*}{Component}} & \multicolumn{2}{c}{CUB} & \multicolumn{2}{c}{SUN} & \multicolumn{2}{c}{AWA2} \\
                          \cmidrule(r){2-3}  \cmidrule(r){4-5}  \cmidrule(r){6-7}
                                 & T1         & Hm         & T1         & Hm         & T1          & Hm         \\
\midrule
Original DAZLE (baseline) & 54.2 $\pm$ 1.7 & 51.0 $\pm$ 0.7 & 60.3 $\pm$ 0.2 & 34.2 $\pm$ 0.01 & 68.9 $\pm$ 1.0 & 70.9 $\pm$ 0.6 \\
Backbone change to T3     & 60.5 $\pm$ 1.6 & 57.1 $\pm$ 0.7 & 57.1 $\pm$ 0.5 & 30.7 $\pm$ 0.1 & 61.5 $\pm$ 0.7 & 62.7 $\pm$ 0.3 \\
Fictitious classes        & 64.1 $\pm$ 1.4 & 59.8 $\pm$ 0.7 & 58.6 $\pm$ 0.6 & 30.8 $\pm$ 0.1 & 65.5 $\pm$ 0.8 & 65.7 $\pm$ 0.2 \\
End-to-End training       & 72.6 $\pm$ 0.5 & 69.9 $\pm$ 0.3 & 59.0 $\pm$ 0.8 & 32.5 $\pm$ 0.2 & 66.4 $\pm$ 0.6 & 67.2 $\pm$ 0.3\\
\bottomrule
\end{tabular}

}
\end{center}
\end{table}

We conduct an ablation study to determine the contribution of each component/architectural change to the overall performance. We conduct the experiments on all three datasets using the three validation folds as defined for each dataset by \cite{xian2017zero}. Report of the average performance on the 3 folds is presented in Table~\ref{table_ablation}. 

Changing the backbone from ResNet-101-L4 to DenseNet-201-T3 gives a boost with 7\% increase in Hm and 7\% increase in T1, but it decreases performance in both SUN and AWA2 (~3-4\% in SUN and ~7-8\% in AWA2).
This leads us to recommend that backbones must be cross-validated for each dataset to achieve optimal performance.

Exposing the model to fictitious classes during training gives an increase of 2\% in Hm and 4\% in T1 in the CUB dataset, little to no improvement in the SUN dataset (0.2\% in Hm and 1.5\% in T1), while in AWA2 we get 3\% increase in Hm and 4\% increase in T1. This leads us to conclude that not all datasets would benefit greatly from adding fictitious classes via dropout to training, and it might have to do with the large number of classes in the SUN dataset.

And the last component, which is the end-to-end training, leads to an increase of an additional 10\% in Hm and 8\% in T1 in CUB, little to no improvement in the SUN dataset (1.6\% in Hm and 0.3\% in T1), and in AWA2 we get 1.5\% increase in Hm and 0.9\% increase in T1.
While end-to-end training seems to be the greatest contributor to the performance in the CUB dataset, we note that this is not always the case; in some cases, its benefits might be nonexistent (as seen in SUN) and it can even degrade performance as we have discussed in catastrophic forgetting (Section~\ref{Catastrophic forgetting}).

\section{Conclusions}
\label{conclusions}
In this work, we presented a novel mechanism that is applicable to any discriminative ZSL model. The idea is to expand the set of seen classes to include fictitious classes that are obtained through dropout. Unlike regular dropout regularization, the new method maps the augmented sample to a corresponding new fictitious class. applying our technique on the DAZLE model achieves state-of-the-art performance on the CUB dataset in both the GZSL and CZSL metrics while maintaining  comparable performance against the rest of the benchmarks. We also show that some backbone architectures might not be suitable as the basis for an end-to-end zero-shot learner due to  catastrophic forgetting. 
Promising future research directions include exploring other augmentations or self supervised techniques to generate different types of fictitious classes. It remain to better understand the role of catastrophic forgetting in zero-shot learning. It might be beneficial to investigate this issue in the relation to lifelong learning\cite{kemker2018measuring} and novelty detection \cite{golan2018deep}.

\nocite{*}
\bibliographystyle{vancouver}
\bibliography{Bibliography}

\end{document}


\maketitle

\makeatletter
\setlength{\@fptop}{5pt}
\makeatother
 
\section{Changes to DAZLE}
\label{changes_to_DAZLE}

Our architectural changes to DAZLE are:
\begin{itemize}
    \item We investigate different backbones and different layers for conv-map extraction, We hypothesize that it might be beneficial to use lower convolutional layers instead of the last convolutional layer as lower layers might retain more general low level features in contrast to higher layers where the network starts to discard image differences. Most networks are trained with strong augmentations that encourage the network to discard low level features such as color (with color jitter), however, they might be beneficial to the knowledge transfer and generalizability to the out-of-distribution classes. Also in higher layers, the network begins the mapping to the semantic space of the 1000 Image-Net classes, which might not be relevant to the classes in our ZSL task at hand.

    \item We abandon the calibration loss, since it requires knowledge of the targeted unseen classes during training, which obviously  limit's the method's scalability and applicability to real world scenarios. Instead, we use the calibrated stacking trick to eliminate or at least reduce the inherent bias of the learner toward seen classes.

\end{itemize}

\section{Backbone choice}
\label{Backbone choice}

We conducted experiments on both ResNet and DenseNet determine from which convolution block it is best  to extract features. Experiments were conducted on the CUB three-fold validation dataset, using DAZLE model without any fictitious classes. We also removed the calibration loss and replaced it with calibrated stacking.

The results can be seen in Table \ref{table:backbone}, and they support our hypothesis that intermediate upper layers perform slightly better than the final layers, at least for the CUB dataset.

\section{Hyperparameters}
\label{hyperparameters}
For each experiment that uses fictitious classes we run a hyperparameter search on the validation set and choose the parameters that maximize the average harmonic mean on the three folds.
Table \ref{table:hp_search} shows our hyper parameters.

\begin{table}[ht!]
\centering
\begin{tabular}{@{}ccccc@{}}

\toprule
Backbone          & T1      & $acc_u$ & $acc_s$ & H       \\ \midrule
DenseNet-201-T1 & 21.62\% & 16.58\%    & 21.61\%   & 18.75\% \\
DenseNet-201-D2 & 33.79\% & 27.73\%    & 34.18\%   & 30.61\% \\
DenseNet-201-T2 & 44.19\% & 37.12\%    & 47.61\%   & 41.70\% \\
DenseNet-201-D3 & 54.49\% & 46.48\%    & 57.55\%   & 51.41\% \\
DenseNet-201-T3 & 60.54\% & 51.89\%    & 63.70\%   & 57.15\% \\
DenseNet-201-D4 & 60.70\% & 53.43\%    & 61.18\%   & 56.93\% \\
\midrule
ResNet-101-L1   & 17.61\% & 12.96\%    & 16.82\%   & 14.53\% \\
ResNet-101-L2   & 27.44\% & 21.71\%    & 26.91\%   & 24.00\% \\
ResNet-101-L3   & 54.85\% & 47.13\%    & 60.18\%   & 52.86\% \\
ResNet-101-L4   & 54.17\% & 47.03\%    & 55.85\%   & 51.04\% \\ \bottomrule
\end{tabular}
\caption{The effect of the backbone}
\label{table:backbone}
\hspace{1cm}

\centering
    \begin{tabular}{@{}lllllll@{}}
\toprule
Dataset & Backbone & $m$ & $1-p$    & $\gamma$ &  &  \\ \midrule
CUB     & DenseNet-201-T3      & 70                       & 0.5  & 2.58  &  &  \\
CUB     & ResNet-101-L3      & 30                       & 0.5  & 3.139 &  &  \\
CUB     & ResNet-101-L4      & 30                       & 0.5  & 4.53  &  &  \\
\midrule
AWA2    & DenseNet-201-T3      & 8                        & 0.5  & 4.231 &  &  \\
AWA2    & ResNet-101-L3      & 15                       & 0.5  & 8.721 &  &  \\
AWA2    & ResNet-101-L4      & 8                        & 0.25 & 5.457 &  &  \\
\midrule
SUN    & DenseNet-201-T3      & 30                       & 0.85 & 1.086 &  &  \\
SUN    & ResNet-101-L3      & 25                       & 0.75 & 1.508 &  &  \\
SUN    & ResNet-101-L4      & 8                        & 0.75 & 1.594 &  &  \\ \bottomrule
\end{tabular}
\caption{Hyperparameters chosen for our model from the validation sets.}
\label{table:hp_search}

\end{table}